\documentclass[10pt, a4paper, conference, compsocconf]{IEEEtran}
\ifCLASSINFOpdf
\else
\fi
\usepackage{graphicx}
\usepackage{caption}
\usepackage{subcaption}
\usepackage{arabtex}
\usepackage{amssymb,amsmath}
\usepackage{pgfplots} 
\usepackage{textcomp}
\usepackage{url}
\usepackage[symbol]{footmisc}
\usepgfplotslibrary{external} 

\begin{document}
%
\title{Writing Style Invariant Deep Learning Model for Historical Manuscripts Alignment}


\author{\IEEEauthorblockN{Majeed Kassis}
	\IEEEauthorblockA{Computer Science Department \\ Ben-Gurion University of the Negev\\
		Beer-Sheva, Israel\\
		majeek@cs.bgu.ac.il}
	\and
	\IEEEauthorblockN{Jumana Nassour}
	\IEEEauthorblockA{Computer Science Department \\ Ben-Gurion University of the Negev\\
		Beer-Sheva, Israel\\
		jumanan@cs.bgu.ac.il}
	
	\and
	\IEEEauthorblockN{Jihad El-Sana}
	\IEEEauthorblockA{Computer Science Department \\ Ben-Gurion University of the Negev\\
		Beer-Sheva, Israel\\
		el-sana@cs.bgu.ac.il}}

\maketitle

\begin{abstract}
	Historical manuscript alignment is a widely known problem in document analysis. Finding the differences between manuscript editions is mostly done manually. In this paper, we present a writer independent deep learning model which is trained on several writing styles, and able to achieve high detection accuracy when tested on writing styles not present in training data. We test our model using cross validation, each time we train the model on five manuscripts, and test it on the other two manuscripts, never seen in the training data. We've applied cross validation on seven manuscripts, netting 21 different tests, achieving average accuracy of $\%92.17$. We also present a new alignment algorithm based on dynamic sized sliding window, which is able to successfully handle complex cases.
\end{abstract}


%
\IEEEpeerreviewmaketitle

\section{Introduction}
Manuscript alignment is one of the important tasks in historical manuscript research. It aims to determine the similarities and differences between two versions of a given manuscripts, usually written by different writers. Currently, this tiresome and time-consuming procedure is done manually. The dissimilarity between various copies of the same manuscript arose from the copying procedures of the past. Each time a manuscript is copied, which is done manually, scribes often omit, insert, or replace words to adapt the content to different geographical regions. Sometimes, scholars perceive the copied version as their personal copy and they embed their own explanation and notes on the original manuscript into the copied version itself. Contemporary researchers study the differences between different versions of a manuscript and attempt to explain the reasons behind these differences to reveal the original content of the manuscript.

A deep Siamese network based system for aligning two manuscripts was published in~\cite{kassis2017siamese}. The deep Siamese Convolutional neural network assists the system in deciding whether a pair of images contains the same text. The network trains on several pages of the given manuscripts, and assists the system to align the rest of the pages of these two specific manuscripts.

We present our deep neural network and alignment algorithm both in which provide several improvements in comparison to the work presented in~\cite{kassis2017siamese}. Our deep neural network provides better prediction accuracy by up to 5\% in comparison to the deep neural network presented in~\cite{kassis2017siamese}. Our alignment algorithm is also able to handle more complex cases. We present a deep neural network and show its ability to predict correctly pairs written in never seen writing styles. We also introduce a new alignment algorithm using this model, based on a dynamic sliding widow and takes as an input minimum window size. The new sliding window automatically increases its size as needed. The sliding window can, in each iteration, move one or more steps (subwords) forward.

\begin{figure*}[!bt]
	\centering
	\includegraphics[height=1.5cm]{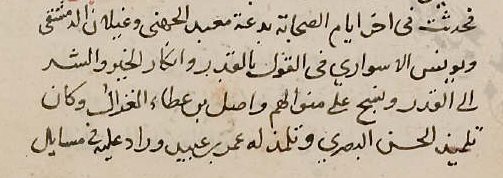}
	\includegraphics[height=1.5cm]{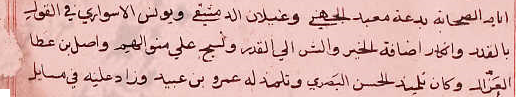}
	\includegraphics[height=1.5cm]{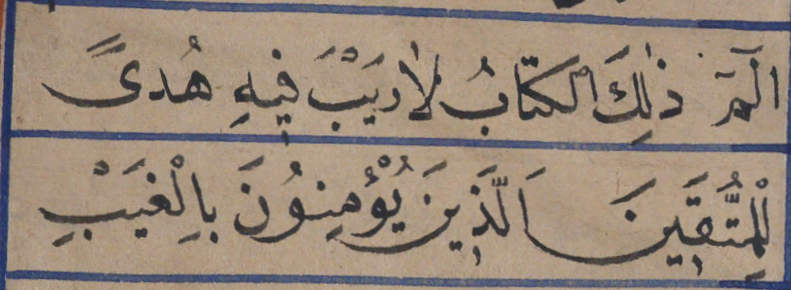}
	\includegraphics[height=1.5cm]{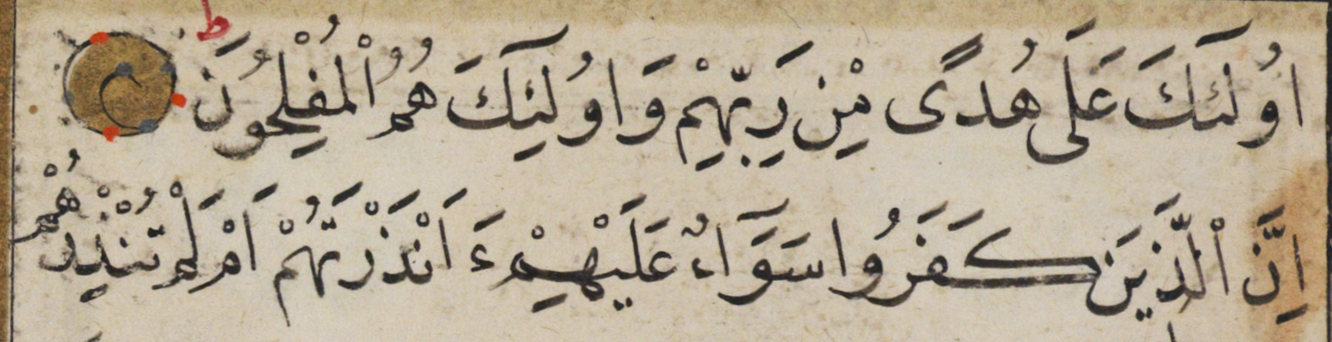}
	\includegraphics[height=1.5cm]{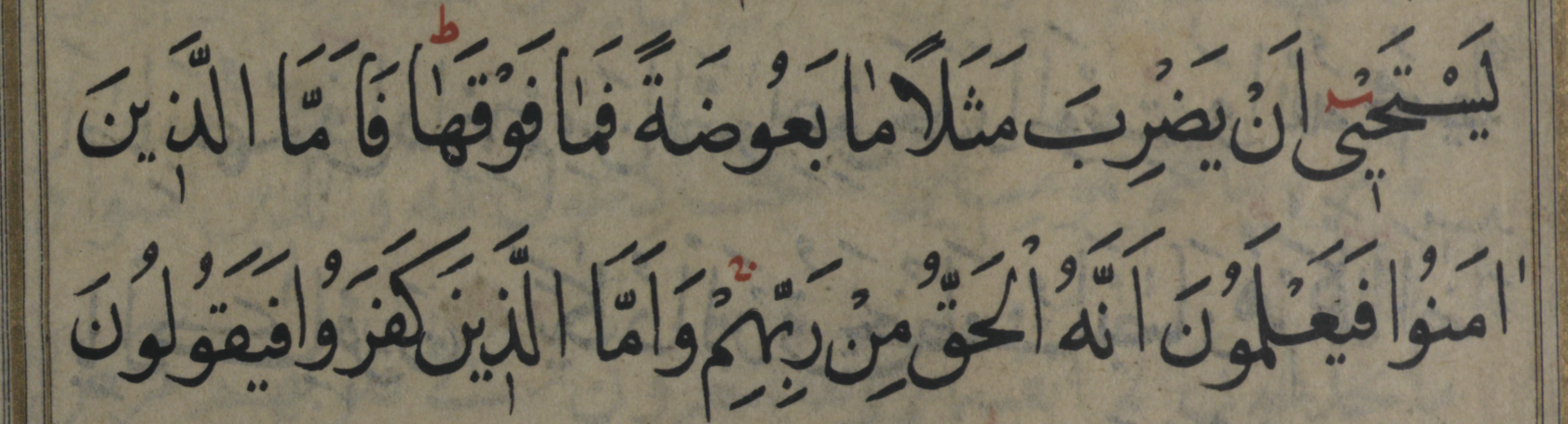}
	\includegraphics[height=1.5cm]{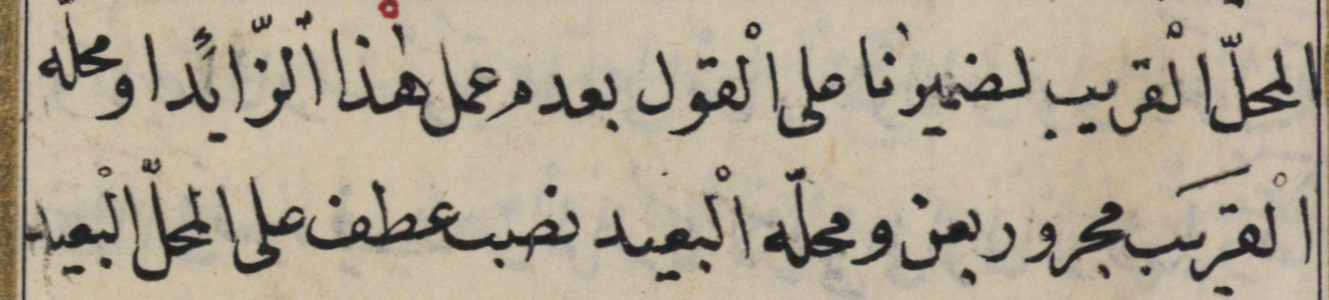}
	\includegraphics[height=1.5cm]{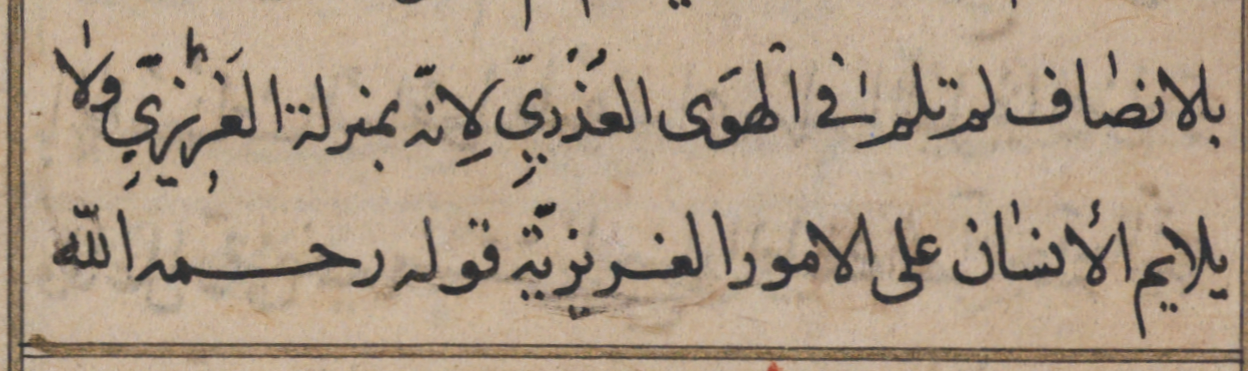}
	\caption{Snippet of the manuscripts with different writing styles. The style difference is illustrated in different stroke widths, and different style in writing certain subwords}
	\label{figure:writingStyles}
\end{figure*}

The rest of te paper is organized as follows. We present related work in Section II, and the system's overview in Section III. In Section IV, we present the model and detail the alignment algorithm, and in Section VI, we show and discuss the results of experiments conducted. Finally, we present our conclusions and planned future work.
 \section{Related Work}

Alignment can be done on several levels, starting from character alignment, word, or sentence alignment. Manuscript alignment can be done in two main ways, supervised, and unsupervised alignment. In supervised alignment, a model is trained, and the alignment system is able to detect similarities using the trained model~\cite{fischer2011transcription, yin2013transcript}. In unsupervised alignment, most works attempt to apply some sort of mapping in some way; aligning images of words with their text transcriptions using some scoring system~\cite{zinger2009text, stamatopoulos2010efficient, stamatopoulos2014novel}. 

Convolutional neural networks has been first presented in~\cite{NIPS1989_293}. In the last decade it began to receive major attention. Every year, new architectures are released~\cite{krizhevsky2012imagenet, zeiler2014visualizing, simonyan2014two, szegedy2017inception}, which was aided by the release of massive datasets such as ImageNet~\cite{russakovsky2015imagenet}.

Due to the shift from feature engineering to feature learning, the interest in convolutional neural networks for handwritten documents has also been growing in recent years. New approaches are introduced every year such as~\cite{sharma2015adapting, jaderberg2014deep, sudholt2016phocnet}.

Siamese Neural Networks were first introduced to solve signature verification as an image matching problem~\cite{bromley1993signature}. A Siamese neural network consists of a network duplicated and joined at its end by a function which decides whether a pair of inputs contain the same text. The parameters between the networks are tied together which makes the network invariant to the order of a pair. Multiple network architectures were presented and used for different tasks such as~\cite{chopra2005learning, koch2015siamese, bertinetto2016fully}.
\section{Overview}
We begin the data preparation step, and the process we initiated to creating the datasets used in training and validation. This data was taken from seven different manuscripts, and using it we generated 21 different versions that consist of a complete cross validation test. Then, we present the deep neural network and the algorithm used in the alignment process. In the alignment process we explain how each case of manuscript differences and the method we used to handle them.
\section{Data Preparation}
In this section we detail the data preparation process. The creation of training, testing, and validation sets, and their structure. The data created will, each time, have samples taken from five manuscripts, while the data created for the validation and test sets will have samples taken from the last two manuscripts only. This process ensures that the model has never seen the writing styles of the testing and validation sets. In total we have generated 21 different version for the cross validation set.

\subsection{Ground Truth Generation}
Recently a new dataset consisting of 668 pages taken from five manuscripts was released~\cite{kassis2017vmlhd}. This dataset is annotated on the subword level and consists of 159,149 subword appearances consisted of 326,289 characters out of a vocabulary of 5,509 forms of sub-words. In addition to this dataset, two more manuscripts were also annotated and added to the dataset, totaling seven manuscripts. An illustration of the writing styles can be seen in Figure~\ref{figure:writingStyles}.

\subsection{Datasets Preparation}
First, we scale the images to fixed width and heights. This process is done by first scaling the images, keeping aspect ratio, to the new fixed height, and then, we resize the images to fit the fixed width, keeping the height intact.

\subsection{Pair Generation}
A pair is called {\em true-pair} if the pair of images contain the same text, and {\em false-pair} if they contain different text. Each dataset has for each subword the same representation ratio in the true-pairs as well as in the false-pairs.

\subsubsection{Training Set}
To apply cross validation, we chose, each time, five manuscripts to generate training samples from, and the last two manuscripts are used to generate samples for validation and testing sets.

To generate true-pairs for the training set, we generate all possible true-pair samples found for each pair of the five manuscripts that were chosen for the testing phase. From these, we randomly chose $400$ samples for each subword form, ensuring that the training set containing at least $20\%$ images taken from each of these manuscripts. There are $1,600$ subword forms on average in each training set.

To generate the false-pairs for the training set, we pair for each image of a sub-word present in the true-pairs sub-set a randomly chosen image containing different text taken from the other manuscripts. This method ensures providing the neural network for each image a true-pair and a false-pair example. This is done for all sub-words present in the true-pairs subset for both manuscripts.

This results in $21$ different training sets for a complete cross-validation (all possible choices of choosing 2 manuscripts for testing out of 7 manuscripts total, leaving the last 5 for training), and range in size between $187,748$ and $203,988$ samples.

\subsubsection{Validation and Test Sets}
To generate both validation and test sets we use the subwords taken the last two manuscripts that were not chosen to be in the training set. This ensures that the network sees completely new data, of different writing styles, not seen in the training phase.

To generate true-pairs for the sets, we look up for each subword its instances in the first manuscript and its corresponding instances in the second manuscript and generate all possible pairs. To generate the false-pairs for the sets, we pair for each image of a sub-word present in the true-pairs subset a randomly chosen image containing different text taken from the other manuscript. This is done for all sub-words present in the true-pairs sub-set for both manuscripts. Once generated, we split the set into two equally sized parts, half of them for the validation set, and the other half for the test set.

The $21$ sets consist of a complete cross-validation test. We generated $21$ training sets, and each of the $21$ sets is split in half, first part  for validation and second part for testing. The smallest set contains $23,948$ samples, and the largest one contains $35,838$ samples. There are $1,600$ subword forms on average in each set.

\section{Manuscript Alignment}
In this section, we detail the deep Siamese neural network in terms of architecture. We also detail the new alignment algorithm explaining the reasoning it. We also details the the accuracy and the robustness of the alignment algorithm, which enabled us to handle complex cases of alignment successfully.

\subsection{Siamese Neural Network}
We base our system on the Siamese Convolutional neural network presented in~\cite{kassis2017siamese} with several modifications that make the network fit better our new data. As can be seen in Table~\ref{table:training} the newly presented deep neural network presents better prediction accuracy. 

The three added Convolutional layers are added between the original Convolutional layers. Each new layer is a hybrid of its two enclosing Convolutional layers. It consists of the same number of features of the previous Convolutional layer, while using the kernel size of the next Convolutional layer.

\begin{figure}
	\includegraphics[width=\linewidth]{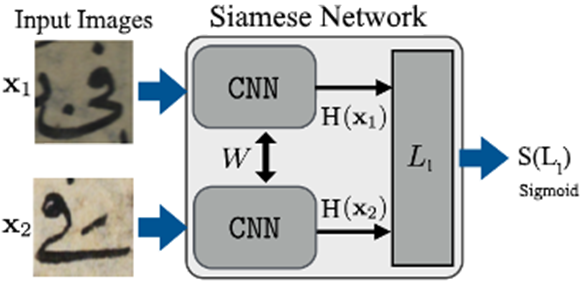}
	\caption{Siamese convolutional neural network architecture. The CNN twin is depicted in its own figure. The main characteristic of the Siamese networks, is that both CNNs share the weights.}
	\label{figure:siamese}
\end{figure}

An illustration of the network architecture can be seen in Table~\ref{table:network}. Finally, we calculate the L1 distance between the two networks, and finalize it by using a sigmoid function. An illustration of the Siamese network can be seen in Figure~\ref{figure:siamese}

\begin{table}
	\centering
	\begin{tabular}{| c | c | c |}
		\hline
		Layer & Filters & Kernel/Pool Size \\ \hline
		Convolution & 64 & 5x5  \\ \hline
		Max Pooling & & 2x2  \\ \hline
		Convolution & 64 & 4x4 \\ \hline
		Convolution & 128 & 4x4  \\ \hline
		Max Pooling & & 2x2  \\ \hline		
		Convolution & 128 & 3x3  \\ \hline
		Convolution & 256 & 3x3  \\ \hline
		Max Pooling & & 2x2  \\ \hline		
		Convolution & 256 & 2x2  \\ \hline
		Convolution & 512 & 2x2  \\ \hline
		Max Pooling & & 2x2  \\ \hline		
		Dense & 4096 & \\ \hline
		DropOut & &  \\ \hline
		Dense & 4096 & \\ \hline
	\end{tabular}
\captionof{table}{CNN architecture, each CNN in the Siamese network consists of the exact architecture detailed here. In every Convolution and Dense layer, we have used ReLU as the activation function. For the DropOut layer we have used a hyperparameter value of 0.1.}
\label{table:network}
\end{table}

\subsection{Manuscript Alignment}

Our alignment system attempts to take in note the cases where manuscripts are different. There are four possible cases of manuscripts alternations as follows: (1) words swapped places with other words, (2) words replaced with other words, (3) words added and do not exist in the other manuscript, (4) word omitted and exists in the other manuscript. Please take note that case (3) and case (4) are exactly the same, and are handled as one case.

We apply the sliding window on both manuscripts we wish to align, and use the model to predict the similarity rating between every pair found in the window. Using the Hungarian method we chose the optimal fit. We do this every iteration until the end of the line. Since data is segmented, the line denotes end of text. Running one iteration provides four possible outcomes, detailed below.

The first possible outcome is identical text of both manuscripts. The criterion are as follows: (1) zero intersections in the pairing result, and (3) all pairs are above similarity rating threshold. As a result, we move the sliding window ahead of step size equal to the sliding window size. 

The second possible outcome is swapping of text. The criterion are as follows: (1) one or more intersections in the result, (2) all pairs are above similarity rating threshold. As a result, we accept the current pairing and acknowledge that subwords have been swapped. We move the sliding window ahead of step size equal to the sliding window size. 

The third possible outcome is added or omitted text to one of the manuscripts. This case criterion are: (1) zero or more intersection in the result, (2) not all pairs are above similarity rating threshold. As a result we increase the window size of the sliding window of the manuscript that contains the added word by one. The sliding window that contains the added word is the sliding window that has the subword matched with another subword found in a farther location along the text line. In this case we increase the sliding window size by one only for the manuscript containing the added word. Then we apply the alignment algorithm once again until one of the sliding windows has all its subwords pairs with above threshold confidence rating.

The fourth, and last possible outcome is a mix of omitted text and a swap of text in the current window. In this case we make sure to solve the case of omitted text by increasing window size of one of the two following the 3rd possible outcome case until it is solved. Once solved we also solve the swapping issue following the second possible outcome case.

In all cases, if we do not pass the requested confidence threshold we move one subword ahead instead, in attempt to further improve the confidence rating by averaging up the results of two consequence sliding windows. Once we achieve the confidence threshold we move ahead a complete window instead of one step ahead.
\section{Experiments}
In this section we will detail the types of experiments we have conducted on the deep Siamese neural network. The main purpose of the experiments is to provide experimental results shows improvements in both prediction and alignment accuracy.

In the previous section, we generated $21$ different tests in total. Each time we train the network on five manuscripts, and keep the last two for testing and validation the results. Since we have seven manuscripts, each has its own writing style, we train the network on five writing styles, and test and validate on the last two writing styles that the network has not seen or trained on. Applying $21$ tests in total, ensures a complete cross-validation of our dataset, and the results illustrates the writing style invariance of the deep neural network.

\subsection{Siamese Neural Network}
After the data generation phase, and the creation of the training, testing, and validation sets. We initialized our network weights. Both the convolutional layers and the fully connected layers are initialized from a normal distribution with zero-mean. The input is a pair of gray-scale images of size $83 \times 69$. 

We tested the architecture on $21$ different training sets, and tested the models on their corresponding validation and test sets. We trained over $200$ epochs for each training set, and chose the model of the epoch providing the highest accuracy on the corresponding validation set. For complete results of these experiments, refer to Table~\ref{table:training}. It is important to note that the first column in the table denotes the two manuscripts chosen in each test for creating the validation and testing sets, which leaves the other five manuscripts to create the training sets.

\newcommand{\specialcell}[2][c]{%
	\begin{tabular}[#1]{@{}c@{}}#2\end{tabular}}

 \begin{table*}[t]
	\centering
	\begin{tabular}{| c | c | c | c | c | c | c | c |}
		\hline
		\specialcell{Test-Validation \\ Manuscripts} & 
		\specialcell{(Ours) Test \\Set Accuracy} & 
		\specialcell{(Ours) Validation \\Set Accuracy} & 
		\specialcell{(Ref~\cite{kassis2017siamese}) Test \\Set Accuracy} & 
		\specialcell{(Ref~\cite{kassis2017siamese}) Validation \\Set Accuracy}  & 
		\specialcell{Training \\Set Size} & 
		\specialcell{Validation Set \\Size} &
		\specialcell{Test Set \\Size} \\ \hline
1,2	&			84.79	\%&	84.44	\%&			80.85	\%&	80.75	\%&	120900	&	27600	&	27600	\\ \hline
1,3	&			89.27	\%&	89.26	\%&			85.36	\%&	85.62	\%&	149800	&	21100	&	21100	\\ \hline
1,4	&			90.57	\%&	90.18	\%&			88.61	\%&	88.41	\%&	149800	&	20600	&	20600	\\ \hline
1,5	&			90.44	\%&	90.43	\%&			88.85	\%&	88.86	\%&	149800	&	20500	&	20500	\\ \hline
1,6	&			89.63	\%&	89.66	\%&			88.26	\%&	87.80	\%&	147500	&	19600	&	19600	\\ \hline
1,7	&			87.13	\%&	87.22	\%&			84.90	\%&	85.30	\%&	147400	&	20800	&	20800	\\ \hline
2,3	&			91.63	\%&	91.72	\%&			89.43	\%&	89.44	\%&	149700	&	21100	&	21100	\\ \hline
2,4	&			92.79	\%&	92.67	\%&			90.32	\%&	90.39	\%&	149700	&	20800	&	20800	\\ \hline
2,5	&			91.77	\%&	91.95	\%&			89.60	\%&	89.57	\%&	149700	&	20500	&	20500	\\ \hline
2,6	&			92.45	\%&	92.29	\%&			89.39	\%&	89.04	\%&	146700	&	19800	&	19800	\\ \hline
2,7	&			91.78	\%&	91.27	\%&			87.97	\%&	88.18	\%&	147600	&	21200	&	21200	\\ \hline
3,4	&			94.00	\%&	93.82	\%&			91.37	\%&	91.77	\%&	152300	&	25400	&	25500	\\ \hline
3,5	&			92.73	\%&	92.61	\%&			90.35	\%&	90.53	\%&	152800	&	25500	&	25600	\\ \hline
3,6	&			93.88	\%&	93.76	\%&			91.32	\%&	91.18	\%&	147900	&	18600	&	18700	\\ \hline
3,7	&			93.45	\%&	93.66	\%&			91.54	\%&	90.98	\%&	147900	&	19000	&	19000	\\ \hline
4,5	&			92.58	\%&	92.70	\%&			91.04	\%&	91.19	\%&	152800	&	25300	&	25400	\\ \hline
4,6	&			95.23	\%&	95.13	\%&			93.57	\%&	93.46	\%&	149100	&	18400	&	18400	\\ \hline
4,7	&			94.28	\%&	94.44	\%&			92.80	\%&	92.72	\%&	180700	&	22600	&	22700	\\ \hline
5,6	&			94.80	\%&	94.71	\%&			93.23	\%&	93.27	\%&	198700	&	27200	&	27300	\\ \hline
5,7	&			93.59	\%&	93.61	\%&			91.15	\%&	91.18	\%&	182200	&	22900	&	22900	\\ \hline
6,7	&			94.55	\%&	94.37	\%&			93.82	\%&	93.45	\%&	198700	&	27200	&	27300	\\ \hline

	\end{tabular}
	\captionof{table}{Accuracy results for validation and test sets of the $21$ cross validation tests. In each test we chose two manuscripts for the test and validation subsets, and used the other five manuscripts for training data. We compare our work with~\cite{kassis2017siamese} as seen in the last two columns. The manuscripts are numbered 1 to 7.}
	\label{table:training}
\end{table*}

\begin{figure}
	\includegraphics[width=0.45\linewidth]{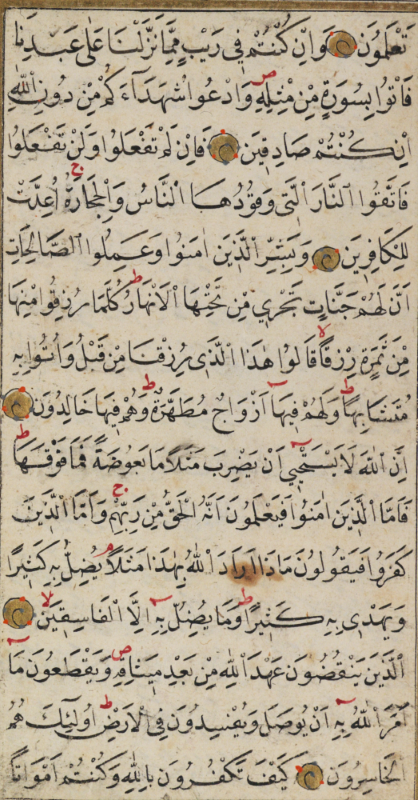}
		\includegraphics[width=0.45\linewidth]{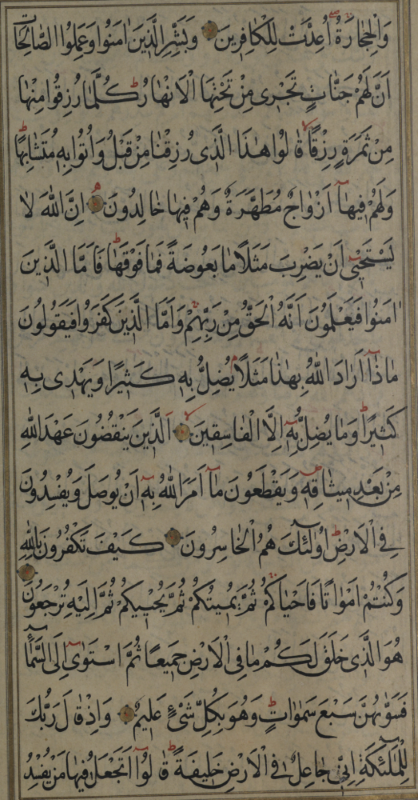}
	\caption{A small snippet of manuscripts number 5 and 6. These manuscripts have an almost identical text.}
	\label{figure:manuscripts56}
\end{figure}

\begin{figure}
	\includegraphics[width=0.45\linewidth]{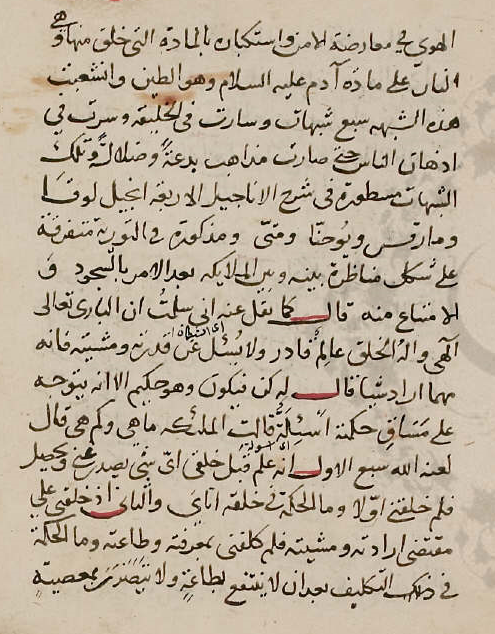}
	\includegraphics[width=0.45\linewidth]{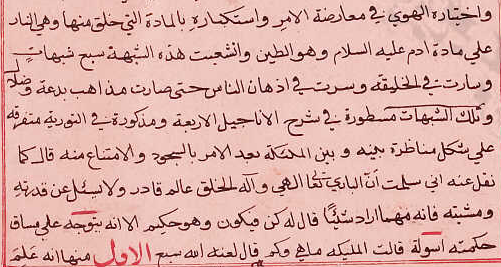}
	\caption{A small snippet of manuscripts number 1 and 2. These manuscripts have more differences in part of their text. Looking at the text you can see that in one place, 4 words were added, and in another place one word was added.}
	\label{figure:manuscripts12}
\end{figure}

\subsection{Manuscript Alignment}
The first test, which is an easy test, due to the fact that the two manuscripts, numbered 5 and 6, as seen in Figure~\ref{figure:manuscripts56}, consist of the same text with minimal differences, over $26$ pages ($52$ pages total), the trained model achieved an accuracy of $92.03\%$ over the validation set. The alignment algorithm achieved $90.46\%$ alignment accuracy using a minimal size $5$ for the sliding window. The second test was conducted on manuscripts 1 and 2, as seen in Figure~\ref{figure:manuscripts12}, of a more challenging nature. These two manuscripts have higher variance in text differences, which consists of word or more swaps, omission (or addition) of one or more words, up to complete sentence, these two manuscripts provide a challenging alignment task. We have used the $36$ annotated pages of each manuscript ($72$ pages total). The trained model achieved an accuracy of $88.59\%$ over the validation set. We've applied the alignment algorithm using a sliding window of minimal size of $13$ subwords, each time applying the alignment algorithm. As a result we were able to align the two manuscripts with alignment accuracy of $85.64\%$. The minimal size was chosen amongst multiple tests and choosing the best result.
\section{Conclusion and Future Work}
In this work we've presented a new method for aligning historical handwritten manuscripts. Using a deep Siamese neural network model, trained on several writing styles, it can predict with high accuracy unseen samples. We showed the results of $21$ tests which depict a full cross validation test done on seven manuscripts, and compared them with related work showing an improvement up to 5\%. We also introduced an alignment algorithm that can align manuscripts consisting of complex cases of text differences.

The planned future work is to train the network on several manuscripts of several languages that share the same script, such as Arabic, Urdu, and Persian and see whether we can train a model on several languages, and attempt to align any script of any language using one model only.
\bibliographystyle{unsrt}
\bibliography{alignment}

\end{document}